\documentclass[letterpaper, 10 pt, conference]{ieeeconf}  % Comment this line out if you need a4paper

\IEEEoverridecommandlockouts                              % This command is only needed if 
\usepackage{graphics} % for pdf, bitmapped graphics files
\usepackage{amsmath} % assumes amsmath package installed
\usepackage{amssymb}  % assumes amsmath package installed
\usepackage{graphicx}
\usepackage[hidelinks]{hyperref}
\usepackage{algorithm}
\usepackage{algorithmic}
\usepackage{booktabs}
\usepackage[table]{xcolor}
\usepackage{comment}

\title{\LARGE \bf
Beyond Noise Steering: Dual-Latent Space Reinforcement\\ Learning for Generative Robot Policy
}

\author{Pengfei Zhang, Teng Sun, and Xianchao Xiu \IEEEmembership{Member,~IEEE}
\thanks{This work was supported by the National Natural Science Foundation of China under Grant 12371306. (\textit{Corresponding author: Xianchao Xiu}.)}
\thanks{Pengfei Zhang, Teng Sun, and Xianchao Xiu are with the School of Mechatronic Engineering and Automation, Shanghai University, Shanghai 200444, China (e-mail:  \texttt{xcxiu@shu.edu.cn}).}
}

\begin{document}

\maketitle
\thispagestyle{empty}
\pagestyle{empty}

%%%%%%%%%%%%%%%%%%%%%%%%%%%%%%%%%%%%%%%%%%%%%%%%%%%%%%%%%%%%%%%%%%%%%%%%%%%%%%%%
\begin{abstract}

Pretrained generative robot policies learn expressive action priors from demonstrations. However, existing reinforcement learning methods only steer the noisy space but fail to modulate intermediate action representations during the generation process, resulting in performance degradation and inefficiency.
To address this limitation, we propose a novel Dual-Latent Space Reinforcement Learning (DLSRL) framework, which complements initial-noise steering with representation-level control inside the frozen generator. Specifically, our actor network predicts two distinct latent variables: an initial-noise latent variable that steers behavior generation, and an action-representation latent variable for intermediate feature modulation. Moreover, this representation latent variable is mapped to adapter features and ingeniously injected into the hidden states of intermediate action tokens via residual connections. Our dual-control design enables direct adjustment of action representations without updating the base policy. Experiments across generative policy architectures and robotic manipulation tasks show that  DLSRL effectively accelerates online robot policy adaptation and achieves competitive performance. Our code is available at \href{https://github.com/xianchaoxiu/DLSRL}{https://github.com/xianchaoxiu/DLSRL}.

\end{abstract}

%%%%%%%%%%%%%%%%%%%%%%%%%%%%%%%%%%%%%%%%%%%%%%%%%%%%%%%%%%%%%%%%%%%%%%%%%%%%%%%%
\section{INTRODUCTION}

Vision-Language-Action (VLA) models provide a scalable route toward general-purpose robotic manipulation by integrating visual perception, language understanding, and action prediction \cite{brohan2022rt1,openx2023,ma2026survey}. Given images, language instructions, and robot states, VLA models can generate actions while leveraging semantic knowledge acquired from large-scale vision-language and robotic datasets \cite{driess2023palme,belkhale2024rt}. Such pretraining reduces the need to learn every behavior from scratch \cite{octo2024,wang2024scaling}. Building on this foundation, recent policy models have also adopted generative action heads to represent continuous and multimodal behavior distributions \cite{shafiullah2022behavior,zhao2023learning,reuss2023goal}. For example, Diffusion Policy \cite{chi2025diffusion} establishes this paradigm for visuomotor control, while RDT-1B \cite{liu2025rdt} and $\pi_0$ \cite{black2025pi0} extend diffusion or flow-matching techniques to large-scale robotic policies. By transforming noise into action sequences, these generative policies provide expressive action priors for complex and dexterous manipulation tasks.

However, when there is a discrepancy with the pretraining distribution, VLA policies often require downstream adaptation \cite{kim2024openvla}. Such deviations are particularly critical in long-horizon tasks, where even small spatial or temporal errors can lead to task failure \cite{ross2010efficient}. Furthermore, this mismatch may only manifest during closed-loop execution and remain absent from small offline adaptation datasets \cite{ross2011reduction}. While full fine-tuning can improve downstream performance, it is computationally expensive for models with billions of parameters. Parameter-efficient methods \cite{houlsby2019parameter,hu2022lora,liu2024dora} reduce the number of trainable parameters, yet they still rely on offline demonstrations and gradient-based policy modifications \cite{kim2025openvlaoft}.  Consequently, online adaptation becomes highly attractive when additional demonstrations are limited, especially when the pretrained policy can remain frozen.

\begin{figure*}[t]
    \centering
    \includegraphics[width=0.98\textwidth]{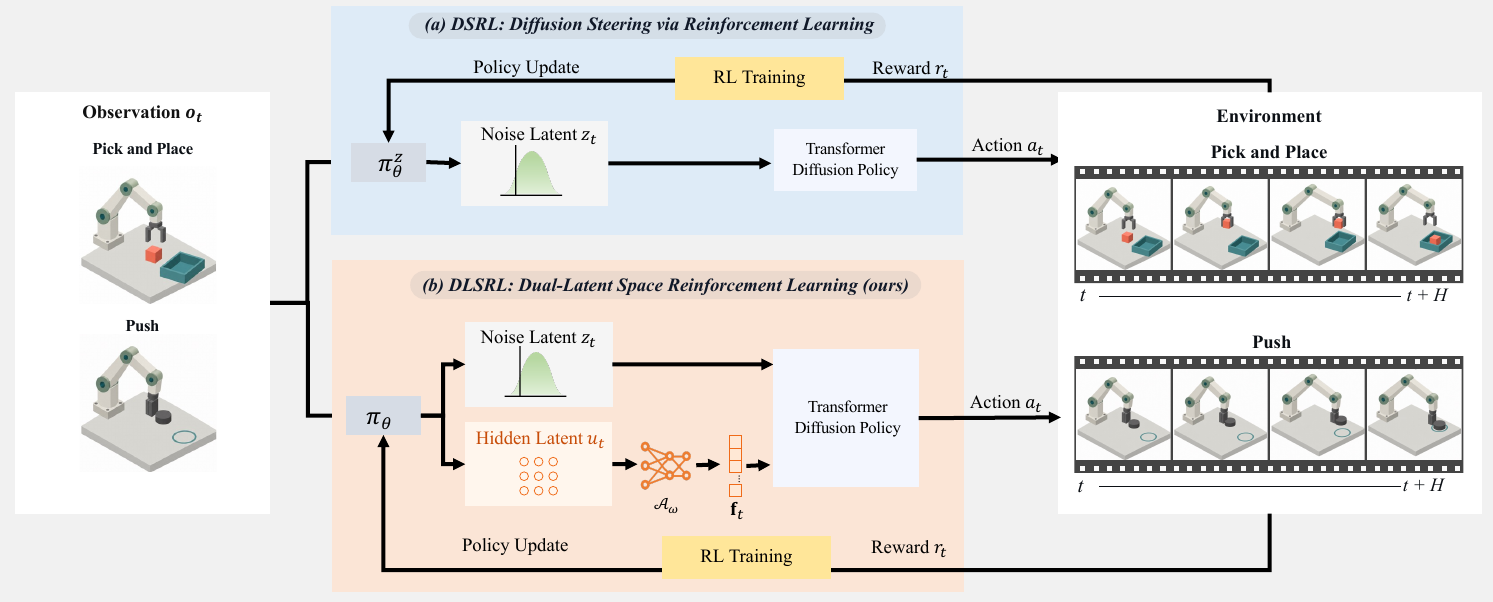}
    \vspace{-0.1cm}
    \caption{Comparison of our DLSRL and existing DSRL, where (a) DSRL learns an observation-conditioned initial-noise latent variable $\mathbf{z}_t$ to steer action generation. 
    (b) DLSRL produces the initial-noise latent variable $\mathbf{z}_t$ and an action-representation latent variable $\mathbf{u}_t$, and the adapter feature mapper $\mathcal{A}_{\omega}$ transforms $\mathbf{u}_t$ into an adapter feature $\mathbf{f}_t$ that modulates the intermediate action-token hidden states. } 
    \label{fig:dlsrl_framework}
\end{figure*}

One popular approach is to improve a generative policy with Reinforcement Learning (RL) \cite{wang2023diffusionql,kang2023efficient}. For instance, Diffusion Policy Policy Optimization (DPPO) directly fine-tunes diffusion policies through policy gradients, showing that generative action models can be optimized online \cite{ren2025dppo}. A complementary line keeps the base policy frozen while learning an external guidance or correction mechanism. For example, Diffusion Steering via Reinforcement Learning (DSRL) learns an observation-conditioned initial-noise policy, thereby steering a frozen generator without backpropagating through its parameters \cite{wagenmaker2025steering}. More broadly, related approaches include using external models to guide denoising trajectories, learning residual actions on top of a fixed policy, or combining human corrections with noise-space optimization \cite{du2025dynaguide,johannink2019residual}. Across these methods, a common characteristic is that they can achieve task-specific adaptation without updating the pretrained generator itself \cite{yuan2025policydecorator,lu2026unisteer}.

More specifically, existing interfaces act mainly at the boundaries of action generation. While initial-noise steering changes the starting point and can influence the behavioral mode selected by the generator, its effect on later representations remains indirect. In contrast, residual policies apply corrections after action generation, thereby failing to shape the evolution of internal action representations \cite{silver2018residual}. This limitation becomes particularly critical when downstream tasks require precise corrections to contact locations, action magnitudes, or local trajectories during generation. As illustrated in controllable image-generation methods, injecting lightweight features into intermediate representations can modulate a frozen generator without relearning its full capabilities \cite{mou2024t2iadapter,zhang2023controlnet,li2023gligen}. However, their control conditions and training objectives do not directly address online robotic feedback. \textit{What is still missing is an RL mechanism that combines global behavior selection with direct modulation of intermediate action representations.}

Motivated by these observations, we propose Dual-Latent Space Reinforcement Learning (DLSRL), which, to the best of our knowledge, is the first work that integrates initial-noise steering with action-token hidden-state modulation. Fig.  \ref{fig:dlsrl_framework} illustrates the difference with the benchmark DSRL \cite{wagenmaker2025steering}. In summary, the main contributions are as follows:
\begin{itemize}
    \item A novel latent space RL framework is introduced by simultaneously optimizing initial-noise latent variables and action-representation latent variables, thereby extending the policy guidance from the sampling initialization phase to the intermediate representation phase.
    \item A lightweight residual modulation mechanism is developed to map the action-representation latent variables into adapter features aligned with action tokens as well as inject them into the Transformer hidden states without requiring updates to the base policy.
    \item A series of experiments on diffusion and flow-matching policies confirm that our proposed DLSRL can improve online adaptation speed while maintaining competitive final performance.
\end{itemize}

\section{RELATED WORK}

\subsection{Generative Robotic Policies}

Generative robot policies characterize multimodal behaviors by modeling conditional action distributions \cite{florence2022implicit}. Among them, Diffusion Policy formulates action-sequence generation as conditional denoising \cite{chi2025diffusion,ho2020denoising}, whereas flow matching learns a continuous transport process from noise to actions, providing an alternative modeling paradigm for efficient generation \cite{lipman2022flow}. Building on these generative formulations, Diffusion and flow matching have also been adopted for action generation in VLA architectures \cite{reuss2024multimodal}, enabling robot actions to be generated from multimodal inputs. For example, RDT \cite{liu2025rdt} employs a diffusion Transformer to model robot actions, while $\pi_0$ \cite{black2025pi0} combines a pretrained vision-language backbone with a flow-matching action expert to generate continuous action sequences based on multimodal inputs. Although these models acquire expressive action priors from large-scale demonstrations, efficient adaptation remains necessary when downstream environments differ from the pretraining distribution.

\subsection{Online Policy Steering}

Generative policies can be adapted either by updating the generator itself or by learning an external steering mechanism \cite{ren2025dppo}. For external steering, existing methods intervene at different stages of action generation. DynaGuide \cite{du2025dynaguide} utilizes guidance from an external dynamics model to modify the diffusion denoising trajectory, whereas DSRL \cite{wagenmaker2025steering} learns an observation-conditioned initial-noise policy for a frozen diffusion policy. Other methods that operate in the latent space include UniSteer \cite{lu2026unisteer} and LPS \cite{im2026latent}. UniSteer converts human corrective actions into noise-space supervision through action-to-noise inversion, while LPS employs an action-space critic to optimize a latent steering policy. These methods act through the initial latent variable, the sampling trajectory, or their associated gradients, without directly modifying intermediate representations inside the action-generation network. Beyond latent-space and sampling-trajectory steering, methods such as Residual RL \cite{johannink2019residual} and Policy Decorator \cite{yuan2025policydecorator} correct the final output of a pretrained policy in the environment action space.However, they modify the generated actions during or after the decoding process rather than directly altering internal action-token representations.
In contrast, our DLSRL retains initial-noise steering while introducing a second RL control interface into the intermediate action-token hidden states, with the base generative policy remaining frozen throughout adaptation.

\subsection{Intermediate-Representation Adapters}

Parameter-efficient adaptation transfers pretrained models by optimizing a small number of additional or low-rank parameters while freezing most model weights \cite{houlsby2019parameter,hu2022lora,li2021prefix}. Within this adaptation paradigm, ControlNet \cite{zhang2023controlnet} and T2I-Adapter \cite{mou2024t2iadapte} map external conditions into intermediate features to control frozen diffusion models. However, these methods are primarily designed for image generation tasks and typically rely on predefined conditions with offline supervision.In contrast, Our DLSRL extends intermediate-representation control to online robot adaptation. Specifically,It learns a low-dimensional action-representation latent variable from environmental feedback, maps it into action-token-aligned adapter features, and coordinates this representation-level signal with initial-noise control.

\section{PROPOSED METHOD}

\subsection{Problem Formulation} 
Robotic manipulation is formulated as a Markov decision process. At time step $t$, the policy receives an observation $o_t=(I_t,q_t,\ell)$, where $I_t$, $q_t$, and $\ell$ denote the visual observation, robot proprioceptive state, and language instruction, respectively. Note that the language instruction $\ell$ is omitted for tasks without language conditioning. Given the current observation, the policy generates a continuous action chunk
$\mathbf{a}_t\in\mathbb{R}^{H\times d_a}$, where $H$ is the action prediction horizon and $d_a$ is the dimensionality of each individual action.

We assume access to a generative base policy $G_\phi$ pretrained on offline demonstrations, with its parameters $\phi$ kept frozen. The policy takes random noise $\mathbf{z}_t
\sim \mathcal{N}(\mathbf{0},\mathbf{I})$ as the initial condition of the generation process 
and transforms it into an action chunk conditioned on the current observation, which is formulated as
\begin{equation}
    \mathbf{a}_t
    =
    G_\phi(o_t,\mathbf{z}_t).
    \label{eq:generative_policy}
\end{equation}
Clearly, the above formulation applies to both diffusion and flow-matching policies, as both can be viewed as conditional generative processes that transform initial noise into continuous action chunks.

Our goal is to learn  a lightweight external control module through online interaction with the environment, while keeping the base policy $G_\phi$ completely frozen. The corresponding expected discounted return is given by
\begin{equation}
    J
    =
    \mathbb{E}
    \left[
        \sum_{t=0}^{T-1}
        \gamma^t r_t
    \right],
    \label{eq:objective}
\end{equation}
where $T$ is the interaction horizon, $r_t$ is the task reward obtained at time step $t$, and $\gamma\in [0,1)$ is the discount factor. This setup preserves the actions previously learned by the base policy while avoiding end-to-end online fine-tuning of large generative models.

\subsection{Dual-Latent Actor}

Let's recall existing noise-space RL methods, which employ an actor to generate task-relevant initial noise based on the current observation, i.e.,
\begin{equation}
    \mathbf{z}_t
    \sim
    \pi_\theta^z(\cdot\mid o_t),
    \label{eq:noise_space_sampling}
\end{equation}
where $\pi_\theta^z$ denotes the noise-space policy parameterized by $\theta$.  Subsequently, a frozen base policy utilizes the selected initial noise to generate action segments via Eq. \ref{eq:generative_policy}.
% \begin{equation}
%     \mathbf{a}_t
%     =
%     G_\phi(o_t,\mathbf{z}_t).
%     \label{eq:noise_space_policy}
% \end{equation}
Unlike random sampling from a standard Gaussian distribution, this actor selects $\mathbf{z}_t$ based on environmental observations, thereby steering the frozen base policy to generate actions that better align with the requirements of the current task. However, this approach controls only the initial conditions and does not directly modify the intermediate action representations within the generative network.

\begin{figure}[t]
    \centering
    \includegraphics[width=0.85\linewidth]{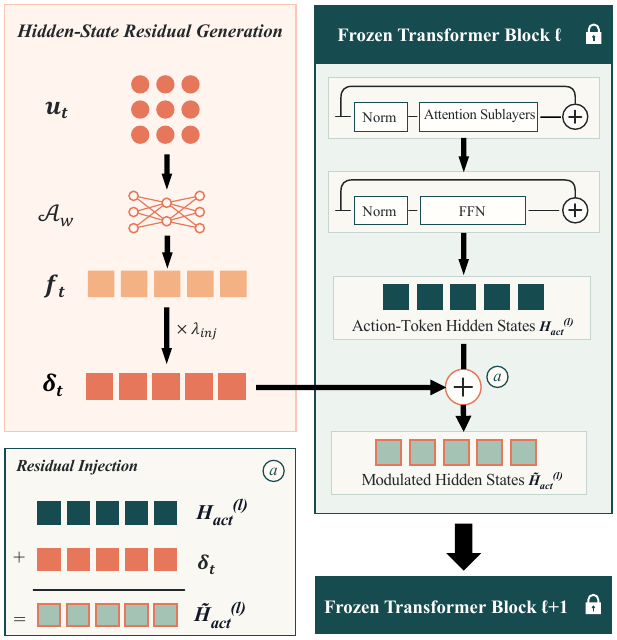}
     \vspace{-0.1cm}
    \caption{Action-token hidden-state modulation in DLSRL, where the adapter feature $\mathbf{f}_t=\mathcal{A}_{\omega}(\mathbf{u}_t)$ is scaled by $\lambda_{\mathrm{inj}}$ and residually injected into the hidden states of a frozen Transformer block.}
    \label{fig:representation_injection}
\end{figure}

Our DLSRL extends the noise-space actor into a dual-latent actor, thereby enabling control that goes beyond initial conditions. Given the current observation $o_t$, the dual-latent actor  produces an initial-noise latent variable $\mathbf{z}_t$ and an action-representation latent variable $\mathbf{u}_t$ according to 
\begin{equation}
    (\mathbf{z}_t,\mathbf{u}_t)
    \sim
    \pi_\theta (\cdot\mid o_t),
    \label{eq:dual_latent_policy}
\end{equation}
where $\pi_\theta$ represents the dual-latent space policy. Then, the action-representation latent variable is  mapped to adapter features via a lightweight adapter feature mapper as
\begin{equation}
    \mathbf{f}_t
    =
    \mathcal{A}_\omega(\mathbf{u}_t),
    \label{eq:adapter_feature_mapping}
\end{equation}
where $\mathcal{A}_\omega$ denotes the adapter feature mapper parameterized by $\omega$. In addition, the marginal distributions of the joint policy over $\mathbf{z}_t$ and $\mathbf{u}_t$ are denoted by $\pi_\theta^z$ and $\pi_\theta^u$, respectively.

\begin{algorithm*}[t]
\caption{Dual-Latent Policy Optimization}
\label{alg:dlsrl}
\begin{algorithmic}[1]
\REQUIRE Base policy $G_\phi$, environment $\mathcal{M}$,
injection strength $\lambda_{\mathrm{inj}}$,
environment-step budget $T$, updates per iteration $K$.

\STATE Initialize buffer $\mathcal{D}$,
critics $Q_\psi^{\mathrm{act}}$, $Q_\nu^{\mathrm{lat}}$,
actor $\pi_\theta$, mapper $\mathcal{A}_\omega$.

\STATE Reset $\mathcal{M}$ to obtain $o_0$.
Set $t\leftarrow 0$.

\WHILE{environment-step budget $T$ is not exhausted}

    \STATE Sample $(\mathbf{z}_t,\mathbf{u}_t)$
    using Eq. \eqref{eq:dual_latent_policy}.

    \STATE Compute $\mathbf{f}_t$, $\mathbf{a}_t$
    using Eq. \eqref{eq:adapter_feature_mapping} and Eq. 
    \eqref{eq:dual_latent_generation}.

    \STATE Execute actions, obtain $(r_t,o_{t+1},d_t)$
    and count environment steps.

    \STATE Store $(o_t,\mathbf{a}_t,r_t,o_{t+1},d_t)$
    in $\mathcal{D}$.

    \FOR{$j=1,\ldots,K$}

        \STATE Sample a minibatch from $\mathcal{D}$ and
        update $\psi$ by off-policy Temporal-Difference (TD) learning.

        \STATE Sample
        $\hat{\mathbf{z}}\sim\mathcal{N}(\mathbf{0},\mathbf{I})$ and
        $\hat{\mathbf{u}}\sim\pi_\theta^u(\cdot\mid o)$.

        \STATE Compute $\hat{\mathbf{f}}$, $\hat{\mathbf{a}}$
        using Eq. \eqref{eq:adapter_feature_mapping} and
        Eq. \eqref{eq:dual_latent_generation}.

        \STATE Update only $\nu$ using
        Eq. \eqref{eq:dual_latent_distillation}.

        \STATE Resample $(\mathbf{z},\mathbf{u})$
        using Eq. \eqref{eq:dual_latent_policy} and
        compute $\mathbf{f}$ using
        Eq. \eqref{eq:adapter_feature_mapping}.

        \STATE Update only $(\theta,\omega)$ using
        Eq. \eqref{eq:dual_latent_actor_loss}.

    \ENDFOR

    \STATE If the episode ends, reset $\mathcal{M}$
    to obtain a new $o_{t+1}$.
    \STATE Set $t\leftarrow t+1$.

\ENDWHILE
\end{algorithmic}
\end{algorithm*}

Therefore, the frozen base policy generates an action chunk, which  is defined as
\begin{equation}
    \mathbf{a}_t = G_\phi \left( o_t,\mathbf{z}_t,\mathbf{f}_t \right).
    \label{eq:dual_latent_generation}
\end{equation}
It is worth pointing out that the initial-noise latent variable $\mathbf{z}_t$ controls the starting point of action generation and primarily influences the overall behavioral mode produced by the base policy. In contrast, the action-representation latent variable $\mathbf{u}_t$ modulates intermediate action representations through the adapter feature $\mathbf{f}_t$, providing more direct representation-level control during generation. Both latent variables are optimized using online task rewards. Throughout this process, the base-policy parameters $\phi$ remain frozen, while the dual-latent actor parameters $\theta$ and adapter feature mapping parameters $\omega$ are updated.

\subsection{Action-Token Hidden-State Modulation}

Consider that the action-representation latent variable influences action generation through the intermediate hidden states of the generation network. Our DLSRL does not introduce additional trainable Transformer layers into the base policy. As illustrated in Fig. \ref{fig:representation_injection}, the adapter feature is scaled by $\lambda_{\mathrm{inj}}$ and added to the action-token hidden states at the outputs of selected frozen Transformer blocks. At the $k$-th generation update step, the action-token hidden states produced by the $l$-th frozen Transformer block are given by
\begin{equation}
    \mathbf{H}_{t,k}^{(l)}
    =
    \mathcal{T}_{\phi}^{(l)}
    \left(
        \widetilde{\mathbf{H}}_{t,k}^{(l-1)},
        \mathbf{e}_t,k
    \right),
    \label{eq:frozen_transformer_block}
\end{equation}
where $\mathcal{T}_{\phi}^{(l)}$ denotes the $l$-th Transformer block with frozen parameters, $\widetilde{\mathbf{H}}_{t,k}^{(l-1)}$ denotes the modulated action-token hidden states from the preceding block, and $\mathbf{e}_t$  represents the conditional representation derived from the current observation. For diffusion policies, $k$ indexes the reverse denoising steps, whereas for flow-matching policies, $k$ indexes the discrete integration steps.

The action-token hidden states satisfy
$\mathbf{H}_{t,k}^{(l)}\in\mathbb{R}^{N_a\times d_h}$, where $N_a$ is the number of action tokens and $d_h$ is the Transformer hidden dimension. The adapter feature obtained from Eq. \eqref{eq:adapter_feature_mapping} has the same shape, i.e., $\mathbf{f}_t\in\mathbb{R}^{N_a\times d_h}$. For a selected Transformer block $l$, our DLSRL scales the adapter feature and
residually injects it into the output action-token hidden states, yielding
\begin{equation}
\widetilde{\mathbf{H}}_{t,k}^{(l)}
=\mathbf{H}_{t,k}^{(l)}+\boldsymbol{\delta}_t.
\label{eq:hidden_state_injection}
\end{equation}
Here, $\boldsymbol{\delta}_t = \lambda_{\mathrm{inj}}\mathbf{f}_t$
denotes the adapter modulation, obtained by scaling the adapter
feature with the global injection strength $\lambda_{\mathrm{inj}}$. The modulated hidden states $\widetilde{\mathbf{H}}_{t,k}^{(l)}$ are subsequently passed to the next Transformer block. Since $\boldsymbol{\delta}_t\in\mathbb{R}^{N_a\times d_h}$ is already aligned with the action-token hidden states, no additional projection layers or trainable parameters are required within the base policy.

Furthermore, the same adapter feature $\mathbf{f}_t$ is shared across the selected Transformer blocks and generation update steps. Residual injection is applied only to action-token positions and does not directly modify visual, language, proprioceptive, or other contextual tokens. Consequently, our DLSRL modifies intermediate action representations without changing the base-network architecture or updating its parameters.

\begin{figure*}[t]
    \centering
    \includegraphics[width=0.98\textwidth]{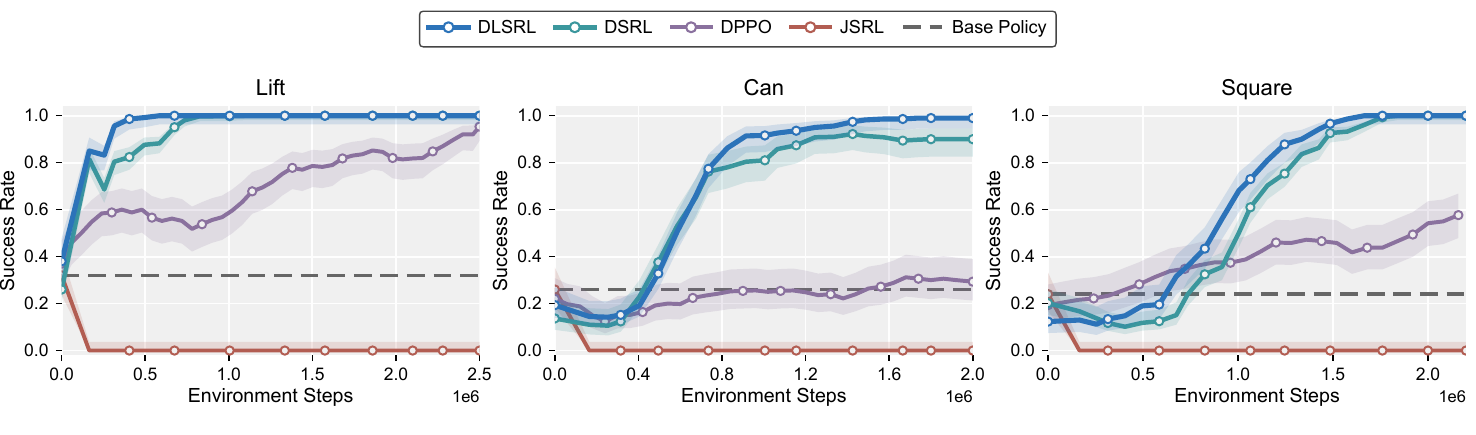}
    \vspace{-0.5cm}
    \caption{Online evaluation success rates of our DLSRL and compared methods on the RoboMimic Lift, Can, and Square tasks, where the horizontal dashed line denotes the performance of the frozen base policy.}
    \label{fig:robomimic_results}
\end{figure*}

\begin{figure*}[t]
    \centering
    \includegraphics[width=0.98\textwidth]
    {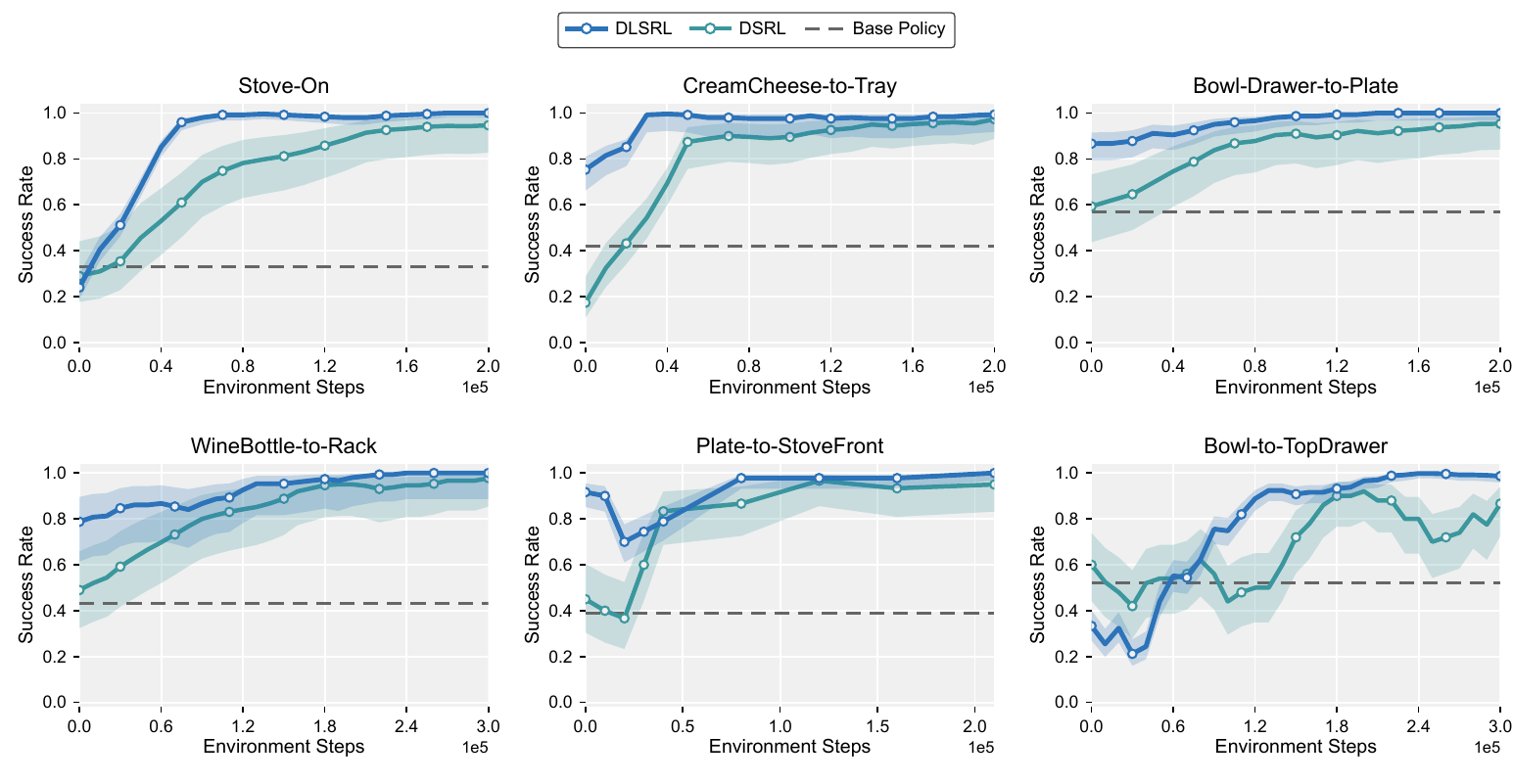}
    \vspace{-0.5cm}
    \caption{Online evaluation success rates of our DLSRL and DSRL  across six LIBERO tasks, where the horizontal dashed line denotes the performance of the frozen base policy.}
    \label{fig:libero_success}
\end{figure*}

\subsection{Dual-Latent Policy Optimization}

Following  the action-to-latent value-distillation mechanism  \cite{wagenmaker2025steering}, we develop a dual-latent policy optimization. This framework contains an action-space critic and a latent space critic, both of which are used only during training. The action-space critic
$Q^{\mathrm{act}}_{\psi}(o_t,\mathbf{a}_t)$ is trained from online interaction data using a standard off-policy temporal-difference objective. It estimates the expected return of action chunks decoded by the frozen base policy under dual-latent control. Then, these value estimates are subsequently distilled into the latent space critic $Q^{\mathrm{lat}}_{\nu}(o_t,\mathbf{z}_t,\mathbf{f}_t)$.

Specifically, given an observation $o_t\sim\mathcal{D}$ sampled
from the replay buffer, we draw $\hat{\mathbf{z}}_t$ from the
standard Gaussian prior and sample
$\hat{\mathbf{u}}_t\sim\pi_\theta^u(\cdot\mid o_t)$.
The corresponding adapter feature $\hat{\mathbf{f}}_t$ and
action chunk $\hat{\mathbf{a}}_t$ are obtained using
Eq.\eqref{eq:adapter_feature_mapping}
and Eq. \eqref{eq:dual_latent_generation}, respectively.
The latent space critic is optimized using the following
value-distillation objective
\begin{equation}
\mathcal{L}_{\mathrm{distill}}
=
\mathbb{E}
\left[
\left(
Q^{\mathrm{lat}}_{\nu}
(o_t,\hat{\mathbf{z}}_t,\hat{\mathbf{f}}_t)
-
\operatorname{sg}
[
Q^{\mathrm{act}}_{\psi}
(o_t,\hat{\mathbf{a}}_t)
]
\right)^2
\right],
\label{eq:dual_latent_distillation}
\end{equation}
where  $\operatorname{sg}[\cdot]$ denotes the stop-gradient operation. During this distillation step, gradients are applied only to the latent space critic parameters $\nu$. This objective allows the latent space critic to learn the joint effect of initial-noise steering and action-token hidden-state modulation on task returns without propagating gradients through the frozen generation process.

Next, the dual-latent actor is optimized by minimizing the following entropy-regularized objective
\begin{equation}
    \label{eq:dual_latent_actor_loss}
    \mathcal{L}_{\mathrm{actor}}
    =
    \mathbb{E}
    \left[
        \alpha
        \log
        \pi_{\theta}^{z}
        (\mathbf{z}_t\mid o_t)
        -
        Q^{\mathrm{lat}}_{\nu}
        (o_t,\mathbf{z}_t,\mathbf{f}_t)
    \right],
\end{equation}
where $\alpha$ is the entropy-regularization coefficient. Although entropy regularization is applied only to the initial-noise branch, gradients of the latent space value with respect to $\mathbf{z}_t$ and $\mathbf{f}_t$ propagate through both actor branches and the adapter feature mapper. During training, the action-space critic, latent space critic, and dual-latent actor together with the adapter feature mapper are updated alternately, while the parameters of the base generative policy $G_\phi$ remain frozen. The overall training procedure is summarized in Algorithm \ref{alg:dlsrl}. At deployment, only the dual-latent actor and adapter feature mapper are retained, and both critics are discarded.

\begin{comment}
\begin{table*}[t]
    \centering
\caption{Average Episode Length on LIBERO Simulation Tasks, where the best results are labeled in bold.}
 \vspace{-0.1cm}
\label{tab:libero_episode_length}
\renewcommand\arraystretch{1.15}
\begin{tabular}{cc ccc c c}
\toprule
Methods & Stove-On &CreamCheese-to-Tray &Bowl-Drawer-to-Plate &WineBottle-to-Rack  &Plate-to-StoveFront &Bowl-to-TopDrawer\\
\midrule
DSRL  & $88.34 \pm 16.42$ & $203.44 \pm 12.58$ & $125.80 \pm 29.18$  & $113.26 \pm 8.97$  & $101.24 \pm 15.82$ & $226.90 \pm 46.04$ \\

DLSRL & $\mathbf{65.90 \pm 1.12}$ & $\mathbf{157.20 \pm 5.77}$  & $\mathbf{97.06 \pm 0.22}$  & $\mathbf{95.96 \pm 0.65}$ & $\mathbf{80.29 \pm 3.42}$ & $\mathbf{180.16 \pm 11.82}$ \\

 \rowcolor{gray!20}   Reduction  & $25.4\%$ & $22.7\%$  & $22.8\%$  & $15.3\%$ & $20.7\%$ & $20.6\%$  \\
\bottomrule
    \end{tabular}
\end{table*}
\end{comment}

\begin{table*}[t]
    \centering
    \caption{Average episode length on LIBERO simulation tasks,
    where the best results are labeled in bold.}
    \vspace{-0.2cm}
    \label{tab:libero_episode_length}
    \renewcommand{\arraystretch}{1.2}

    \begin{tabular}{lcccccc}
        \toprule
        Methods
        & Stove-On
        & CreamCheese-to-Tray
        & Bowl-Drawer-to-Plate
        & WineBottle-to-Rack
        & Plate-to-StoveFront
        & Bowl-to-TopDrawer \\
        \midrule

        DSRL
        & $88.34 \pm 16.42$
        & $203.44 \pm 12.58$
        & $125.80 \pm 29.18$
        & $113.26 \pm 8.97$
        & $101.24 \pm 15.82$
        & $226.90 \pm 46.04$ \\

        DLSRL
        & $\mathbf{65.90 \pm 1.12}$
        & $\mathbf{157.20 \pm 5.77}$
        & $\mathbf{97.06 \pm 0.22}$
        & $\mathbf{95.96 \pm 0.65}$
        & $\mathbf{80.29 \pm 3.42}$
        & $\mathbf{180.16 \pm 11.82}$ \\

        \rowcolor{gray!20}
        Reduction
        & $25.40\%$
        & $22.73\%$
        & $22.85\%$
        & $15.27\%$
        & $20.69\%$
        & $20.60\%$ \\
        \bottomrule
    \end{tabular}
\end{table*}

\section{Numerical Experiments}

Experiments are conducted on RoboMimic and LIBERO to evaluate the performance of our DLSRL across different generative policy architectures. 
On RoboMimic, the Lift, Can, and Square tasks are used with a pretrained Transformer-based Diffusion Policy whose parameters remain frozen during online adaptation. The policy contains four Transformer blocks with a hidden dimension of $128$ and predicts action chunks of length $4$. Under the same environment interaction budget, our DLSRL is compared with Base Policy, JSRL \cite{uchendu2023jump}, DPPO \cite{ren2025dppo},  and DSRL \cite{wagenmaker2025steering}. 
On LIBERO, six tasks, including Stove-On, CreamCheese-to-Tray, Bowl-Drawer-to-Plate, WineBottle-to-Rack, Plate-to-StoveFront, and Bowl-to-TopDrawer are considered for the flow-matching setting with $\pi_0$. Here, the vision-language backbone and action-generation module remain frozen. Moreover, DSRL is used as the main comparison because it adapts the same base policy through initial-noise steering alone. 
In addition, the success rate is evaluated over 100 episodes during the training process.

\subsection{Can DLSRL Improve Online Adaptation Efficiency?}
\label{subsec:diffusion_policy_results}

Fig. \ref{fig:robomimic_results} shows the results of our DLSRL compared with other methods on the RoboMimic Lift, Can, and Square tasks. Across all three tasks, our DLSRL demonstrates faster performance improvement during the early and intermediate stages of training. On Lift, although DSRL gradually narrows the gap in the later stages, our DLSRL reaches a near-saturated performance level much earlier. On Square, our DLSRL exhibits faster and more stable performance gains.  In particular, on our DLSRL on Can reaches approximately $99\%$ success in the later stage of training, compared with about $90\%$ for DSRL. Under the same interaction budget, DPPO improves more slowly, while JSRL does not consistently outperform Base Policy.

Overall, the main advantage of our DLSRL is faster adaptation rather than uniformly higher final performance. It is more pronounced on Can and Square, where successful execution requires greater action precision, whereas the simpler Lift task eventually allows DSRL to approach a similar success rate. These results indicate that our DLSRL makes more effective use of online interactions when adapting a frozen diffusion policy.

\begin{figure}[t]
    \centering
    \includegraphics[width=0.85\linewidth]
    {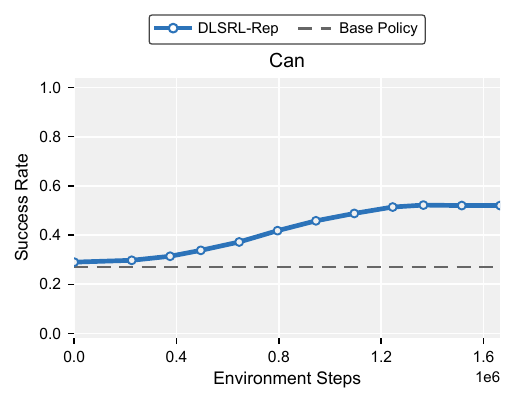}
    \vspace{-0.4cm}
    \caption{Ablation studies on the Can task, where DLSRL-Rep retains only action-token hidden-state modulation without learned initial-noise steering.}
    \label{fig:ablation}
\end{figure}

\subsection{Does DLSRL Transfer to Flow-Matching VLA Policies?}
\label{subsec:pi0_results}

Fig. \ref{fig:libero_success} illustrates our DLSRL with DSRL on six LIBERO tasks using the pretrained $\pi_0$ policy. Across these tasks, our DLSRL generally improves the success rate more rapidly and maintains higher performance over most of the interaction budget. Although both methods continue to improve with training, our DLSRL typically reaches saturated performance with fewer environment interactions.

%The difference is also reflected in task-execution efficiency. 
As listed in Table \ref{tab:libero_episode_length}, our DLSRL achieves shorter average episode lengths on all six tasks, indicating consistent improvements in task execution efficiency across different manipulation scenarios. These consistent improvements demonstrate that the benefit of dual-latent control is not limited to diffusion policies and can also be observed in flow-matching-based VLA policies.

\begin{figure}[t]
    \centering
    \includegraphics[width=0.85\linewidth]
    {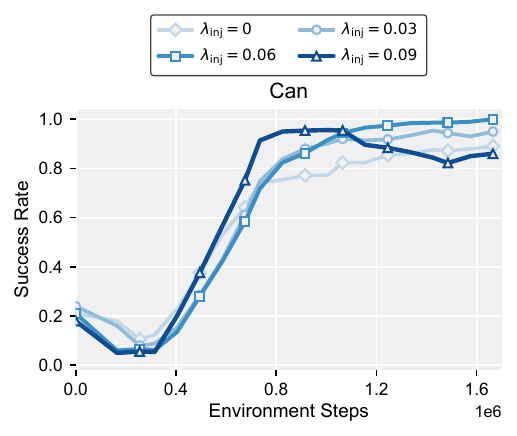}
    \vspace{-0.4cm}
    \caption{Analysis of action-token hidden-state injection strength on the Can task under different values of $\lambda_{\mathrm{inj}}$.}
    \label{fig:injection_scale}
\end{figure}

\subsection{How Does Representation Modulation Help?}

To investigate whether representation-level modulation contributes independently to online adaptation, DLSRL-Rep is evaluated on the Can task with learned initial-noise steering disabled. As shown in Fig. \ref{fig:ablation}, DLSRL-Rep starts from a success rate close to that of the frozen base policy and improves steadily from approximately $27\%$ to $52\%$. This gain  shows that modifying intermediate action-token representations alone can improve the base policy, even without learned initial-noise steering. These results also help explain why combining these two control interfaces in our DLSRL accelerates the adaptation process.

Fig. \ref{fig:injection_scale} analyzes the effect of modulation strength  by varying $\lambda_{\mathrm{inj}}\in\{0,0.03,0.06,0.09\}$ while keeping the remaining training settings fixed. It can be seen  that a larger injection strength does not always lead to better performance. Although $\lambda_{\mathrm{inj}}=0.09$ improves most rapidly at the beginning of training, it exhibits larger fluctuations later on. In contrast, $\lambda_{\mathrm{inj}}=0.03$ produces a more gradual but stable improvement. The intermediate value $\lambda_{\mathrm{inj}}=0.06$ provides the best balance between adaptation speed and stability and eventually reaches a success rate of $100\%$. These observations suggest that effective representation-level control requires sufficient modulation strength without excessively perturbing the pretrained action representations.

\section{Conclusion}
\label{sec:conclusion}

In this paper, we develop an efficient Dual-Latent Space Reinforcement Learning (DLSRL) framework for online adaptation of frozen generative robot policies. Different from the previous work, our DLSRL combines initial-noise steering with lightweight modulation of intermediate action-token representations, providing an additional control interface inside the frozen generator. Experiments on RoboMimic and LIBERO demonstrate that the proposed DLSRL generally improves online adaptation speed and maintaining competitive success rates. 
%Similar gains are observed for both diffusion and flow-matching policies while the pretrained base policy remains frozen. 
Future work will evaluate our DLSRL on real-world environments and investigate the effects of injection layers, generation steps, and injection strengths on adaptation performance.

\bibliographystyle{IEEEtran}
\bibliography{references}

\end{document}